\documentclass[conf]{new-aiaa}
\usepackage[utf8]{inputenc}
\usepackage[T1]{fontenc}
\usepackage{graphicx}
\usepackage{amsmath}
\usepackage[version=4]{mhchem}
\usepackage{siunitx}
\usepackage{longtable,tabularx}
\usepackage{ragged2e}
\usepackage{colortbl}
\usepackage{float}

\title{Assessing the Criticality of Longitudinal Driving Scenarios using Time Series Data}
\author{Nico Schick\footnote{N. Schick, M. Sc. studied Applied Computer Sciences (M. Sc.) and Computer Engineering (B. Eng.) at Esslingen University of Applied Sciences. Email: Nico.Schick@hs-esslingen.de}}

\begin{document}
\maketitle

\begin{abstract}
Unfortunately, many people die in car accidents. To reduce these accidents, cars are equipped with driving safety systems. With autonomous vehicles, the driver's behavior becomes irrelevant as the car drives autonomously. All autonomous driving algorithms must undergo extensive testing and validation, especially for safety-critical scenarios. Therefore, the detection of safety-critical driving scenarios is essential for autonomous vehicles. This publication describes safety indicator metrics based on time series covering longitudinal driving data to detect safety-critical driving scenarios. 
\end{abstract}

\textbf{Scientific Question:} \textit{Which safety indicator metrics can be used to identify safety-critical driving scenarios based on time series covering longitudinal driving data?}


\section{Motivation}
Approximately 1.3 million people die in road crashes each year, often due to self-inflicted mistakes or the misconduct of other drivers \cite{c21}. To reduce road accidents, the automotive industry has introduced the concept of vehicles equipped with advanced driver-assistance systems or the use of autonomous vehicles.

To ensure public trust and road safety, all autonomous driving algorithms must undergo thorough testing and validation, especially for safety-critical driving scenarios that pose risks to humans and the environment. Therefore, the detection of safety-critical driving scenarios is crucial for autonomous vehicles. \cite{pubS1}

Several approaches have been proposed in the literature to identify safety-critical driving scenarios such as hazard analysis techniques for system safety \cite{Sev}. 

This publication describes safety indicator metrics that can be used to identify safety-critical driving scenarios based on time series covering longitudinal driving data. The metrics are evaluated and compared based on defined criteria specific to the use case. One of the most suitable safety indicator metric is examined thoroughly for the underlying use case, employing synthetic data that encompasses a safety-critical driving scenario as a validation example.

\newpage
\section{Time Series}

Time series are sequences of ordered observations of a specific characteristic corresponding to a certain period of time. The time values can be represented as $t = t_1, \ldots, t_N$, with $N$ as the total number of observations.

The time vector can be arranged with constant, regular, or irregular distances. \cite{DIPEIS} In general, when assigning one data point of one size to one time value, it is referred to as a univariate time series. On the other hand, when assigning data points of multiple sizes or vectors to one time value, it is referred to as a multivariate time series.

Distinctions are defined based on continuous and discrete time space. Continuous time series focus on sequential data points without temporal or spatial interruptions, such as sensor values. \cite{BAEU} Discrete time series, on the other hand, refer to data points that correspond to asynchronous time events, such as event-based driving scenarios.

In this publication, multivariate time series are considered since they are also relevant in the real world, taking into account sensor values collected during driving, such as position, speed, or acceleration sensor values, all pertaining to the same underlying time vector.

In general, a multivariate time series in real space can be defined as follows:
\begin{align}
    f : D \in \mathbb{R} \mapsto Z \in \mathbb{R}^k, t \mapsto Y \in  \{Y_{t=t_1}^k, Y_{t=t_2}^k, ..., Y_{t=t_N}^k \}, \#(t,Y)=N,k \in \mathbb{N}^+
\end{align}
The time vector is defined by $t$. The observations $Y$ of the characteristics $k$ are combined in $Y^k$. The number of data points in the multivariate time series is defined by $N$. \cite{BARC}

\section{Autonomous Driving}

\begin{figure}[H]
\centering
    \includegraphics[scale=0.48]{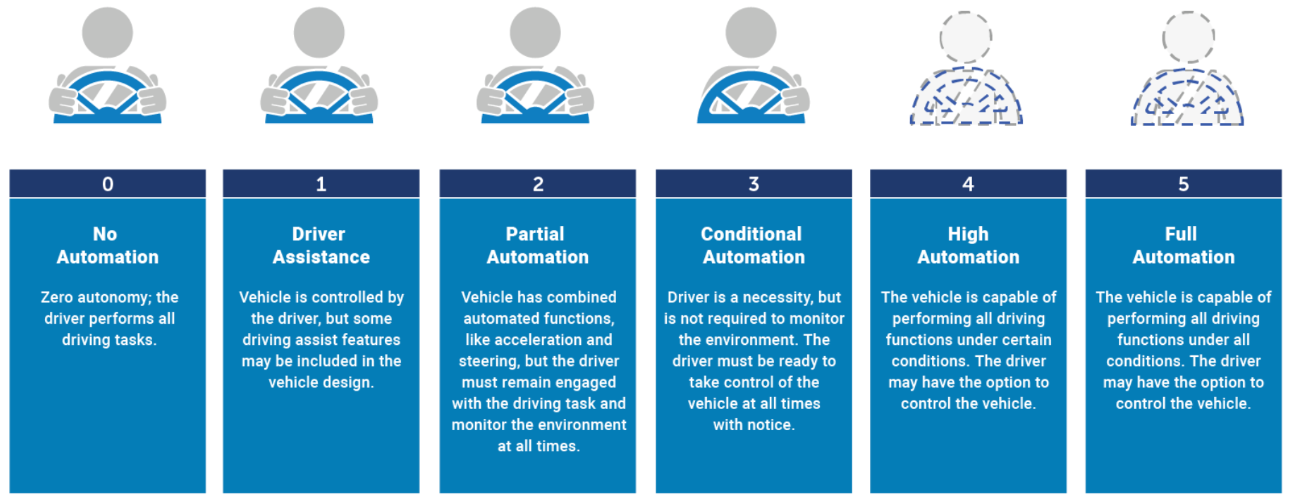}
    \caption{SAE Levels of Autonomy \cite{NAB}}
    \label{fig:sae-levels}
\end{figure}

Vehicles and cars play an essential role in society, both in the private and public sectors. Manufacturers are constantly developing innovative enhancements and improvements. One notable advancement in this regard is autonomous driving. Autonomous vehicles offer increased safety through various assistance systems, also known as advanced driver assistance systems. These systems enable driving without human intervention.  \cite{BARC} The Society of Automotive Engineers (SAE) has defined six levels of driving automation, ranging from Level 0 (fully manual) to Level 5 (fully autonomous) \cite{SYNOPSIS}. Figure \ref{fig:sae-levels} illustrates these levels.

\begin{figure}[ht]
\centering
    \includegraphics[scale=0.55]{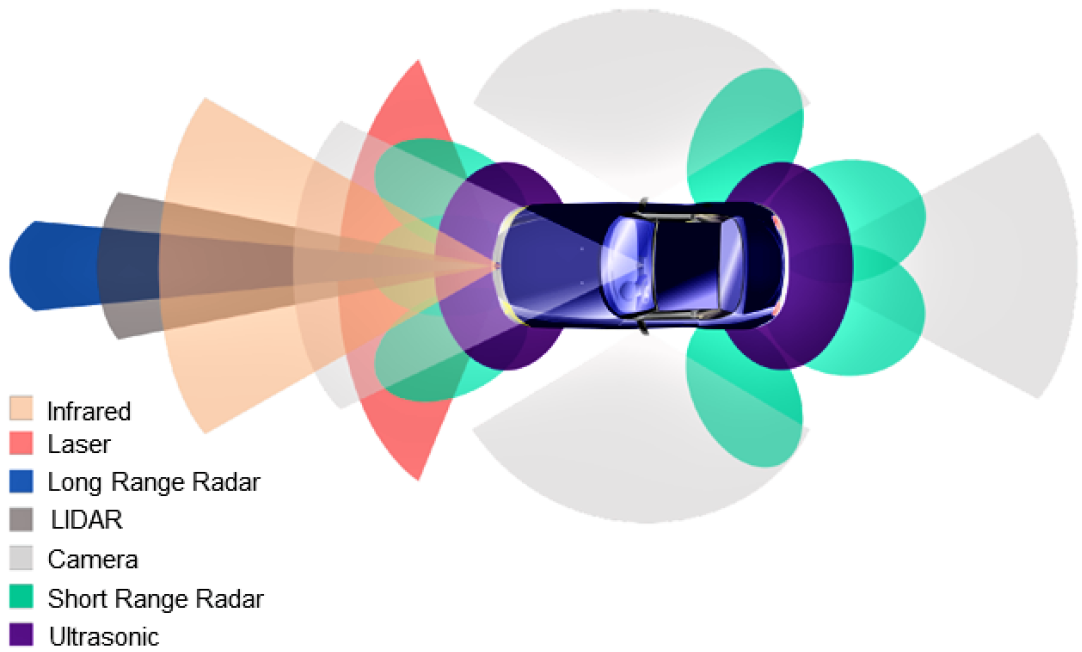}
    \caption{Autonomous Vehicle Sensors \cite{BAMY}}
    \label{fig:sensors}
\end{figure}

Autonomous driving algorithms rely on interpreting information to identify appropriate driving paths and detect obstacles or objects. To achieve this, autonomous cars use a variety of sensors to perceive their surroundings. The main types of sensors are visualized in Figure \ref{fig:sensors}. \cite{KOLLA} \cite{BAMY}

\section{Assessment of Safety-Critical Driving}

\begin{figure}[ht]
\centering
    \includegraphics[scale=0.46]{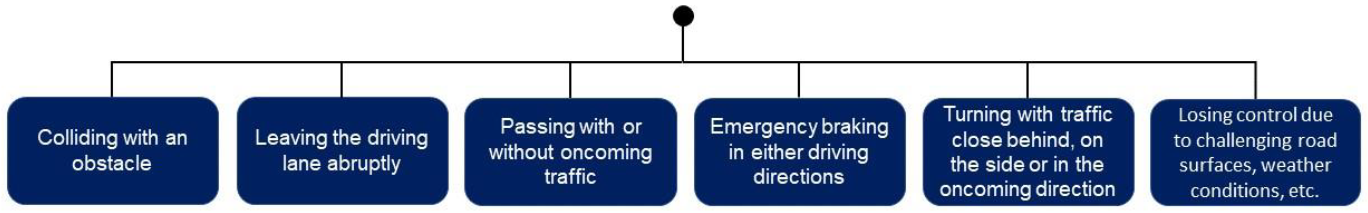}
    \caption{Taxonomy: Safety-Critical Driving Scenarios \cite{BAMY}}
    \label{fig:saf-crit-scenes}
\end{figure}

Safety-critical driving scenarios have the potential to result in severe accidents and loss of human lives. These scenarios can be categorized into different groups. Figure \ref{fig:saf-crit-scenes} presents a taxonomy of safety-critical driving scenarios. \cite{BAMY}

Figure \ref{fig:saf-crit-scenes} provides a generalized taxonomy of these driving scenarios, highlighting the challenge of classifying a driving scenario as safety-critical \cite{BAMY}. For instance, consider the act of pulling out of a parking space. If there are no moving objects nearby, the scenario can be classified as non-safety-critical. However, depending on the distance to the object and its speed, it could become safety-critical. 

The safety relevance of driving scenarios generally depends on factors such as distance, speed differences, and acceleration of the objects involved (e.g., a following drive with two cars). It is crucial to distinguish between safety-critical and non-safety-critical driving scenarios, not only for vehicles in general but also for autonomous vehicles. Various methods can be employed to assess the criticality of driving scenarios. In this publication, we focus on time series data and longitudinal driving, and therefore, we focus on corresponding methods relevant to our specific use case. \cite{BAMY}

\begin{figure}[ht]
\centering
    \includegraphics[scale=0.57]{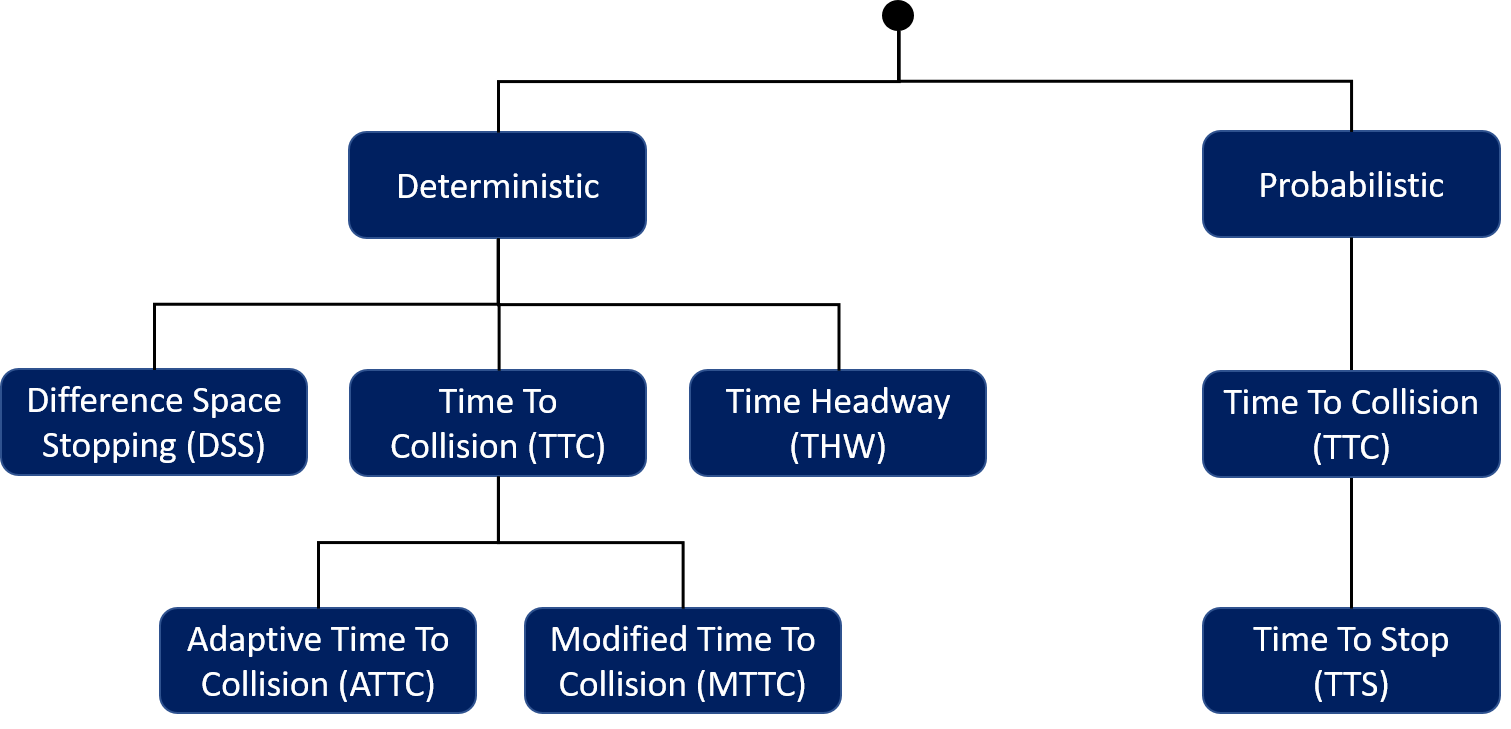}
    \caption{Safety Indicator Metrics (Longitudinal Driving)}
    \label{fig:taxo}
\end{figure}

In fact, Figure \ref{fig:taxo} depicts a taxonomy of safety indicator metrics that can be used to identify safety-critical driving scenarios. It should be noted that \cite{SafeInd1Sum} and \cite{SafeInd2Sum} also cover many different metrics in this regard.

In general, the risk assessment of driving scenarios can be approached deterministically or probabilistically. A deterministic approach relies on rule-based systems that estimate the possibility of collisions as a binary prediction. This approach allows for a clear differentiation between non-safety-critical and safety-critical driving scenarios and depends on certain threshold values. The main advantages of such rule-based systems are simplicity and computational efficiency. However, deterministic approaches do not explicitly model uncertainties, such as transitions between non-safety-critical and safety-critical scenarios. To address uncertainties, probabilistic methods can be employed, which rely on probabilistic descriptions of temporal and spatial relationships between vehicles. These methods can be implemented using various approaches, such as Markov processes, Bayesian networks, Monte Carlo simulations, or fuzzy logic. It should be noted that deterministic and probabilistic methods can also be combined in hybrid approaches. \cite{BAMY}

In the following sections, we provide a detailed description of all the metrics shown in Figure \ref{fig:taxo}. 

\newpage

\subsection{Time To Collision (TTC)}
TTC is defined as the time remaining until a collision occurs \cite{BAMY} \cite{ref2}:

\begin{equation}
TTC = \frac{x_{L} - x_{F} - l_{V}}{v_{F} - v_{L}}
\end{equation}

Here, $x$ represents the position, $v$ represents the speed, and $l_V$ represents the length of the vehicle. The subscripts $L$ and $F$ refer to the leading and following vehicles, respectively. \cite{BAMY} \cite{ref2}

TTC is based on the assumption that both the following and leading vehicles are traveling at a constant speed. It does not consider the acceleration or deceleration of either vehicle. TTC is only valid when the speed of the following vehicle is greater than that of the leading vehicle. \cite{BAMY} \cite{ref2}

\subsubsection{Effective Distance}
\begin{figure}[ht]
\centering
    \includegraphics[scale=0.75]{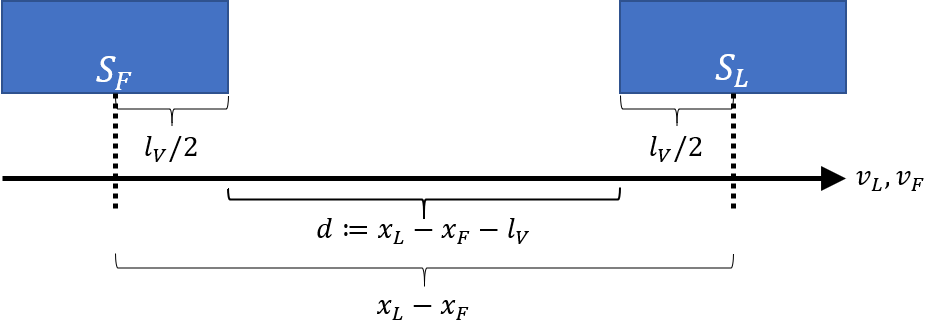}
    \caption{Effective Distance (Longitudinal Driving)}
    \label{fig:vehicle-length}
\end{figure}

Figure \ref{fig:vehicle-length} illustrates the importance of vehicle length for such metrics. The distance between the vehicles depends on the positions of both vehicles as well as the length of the following vehicle, denoted by $l_V$. Consequently, the safety-related distance between the vehicles is defined as $d := x_L - x_F - l_{V}$ (effective distance). In this regard, a homogeneous mass distribution is assumed for both vehicles, with the center of gravity $S$ positioned in the middle.

\subsection{Modified Time To Collision (MTTC)}

MTTC is a modified version of TTC that takes into account changes in the speeds of both vehicles, making it applicable to a wider range of driving scenarios, particularly those involving dynamic behavior of the vehicles. \cite{BAMY} \cite{ref3}

\begin{equation}
MTTC = \frac{-\Delta V \pm \sqrt{\Delta V^2 + 2\Delta A \, d}}{\Delta A}
\end{equation}

Here, $\Delta V$ represents the relative speed ($v_F - v_L$), $\Delta A$ represents the relative acceleration ($a_F - a_L$), and $d$ represents the aforementioned distance ($x_L - x_F - l_{V}$) between the vehicles.

MTTC is considered to be more reliable than TTC because it takes into consideration the dynamic parameters of two consecutive vehicles, including their relative distance, speed, and acceleration, when assessing the likelihood of a conflict occurrence. However, MTTC cannot be used when acceleration values are changing. In summary, MTTC can be applied in car-following scenarios as long as the acceleration values remain constant. \cite{BAMY} \cite{ref3}

\newpage

\subsection{Adaptive Time To Collision (ATTC)}
ATTC is a generalized formulation of TTC based on equations of motion. It can be employed depending on data availability and the required level of accuracy. \cite{BAMY} \cite{ref2} Table \ref{ATTCform} describes the three different cases for calculating TTC \cite{BAMY} \cite{ref2}:

\begin{table}[H]
\begin{tabular}{l||l||l}
\hline
\textbf{TTC Type}&\textbf{Assumptions}&\textbf{Equations}\\
\hline
$TTC_{1}$ &$\dot x_{L}$ = $cte$&$ x_{F} = x_{0F} + \dot x_{F}t$\\
                 &$\dot x_{F}$ = $cte$&$x_{L} = x_{0L} + \dot x_{L}t$\\
                 &$\ddot x_{L}$ = $\ddot x_{F}$ = 0&($\dot x_{F}$ - $\dot x_{L}$) \small$TTC$ - ($x_{L}$ - $x_{F}$ - $l_V$) = 0\\
\hline
$TTC_{2}$ &$\ddot x_{L}$ = $cte$&$x_{F} = x_{0F} + \dot x_{F}t$ + $\frac{1}{2}$$\ddot x_{F}t^2$\\                
		 &$\ddot x_{F}$ = $cte$&$x_{L} = x_{0L} + \dot x_{L}t$ + $\frac{1}{2}$$\ddot x_{L}t^2$\\
	         &$\dddot x_{L}$ = $\dddot x_{F}$ = 0&$\frac{1}{2}$($\ddot x_{F}$ - $\ddot x_{L}$) \small$TTC^2$ + ($\dot x_{F}$ - $\dot x_{L}$) \small$TTC$ - ($x_{L}$ - $x_{F}$ - $l_V$) = 0\\
\hline
$TTC_{3}$ &$\dddot x_{L}$ = $cte$& $x_{F}$ = $x_{0F}$ + $\dot x_{F}t$ + $\frac{1}{2}$$\ddot x_{F}t^2$ + $\frac{1}{6}$$\dddot x_{F}t^3$\\
		 &$\dddot x_{F}$ = $cte$& $x_{L}$ = $x_{0L}$ + $\dot x_{L}t$ + $\frac{1}{2}$$\ddot x_{L}t^2$ + $\frac{1}{6}$$\dddot x_{L}t^3$\\
		 &				 &  $\frac{1}{6}$($\dddot x_{F} - \dddot x_{L}$) \small$TTC^3$ + $\frac{1}{2}$($\ddot x_{F} - \ddot x_{L}$) \small$TTC^2$ + ($\dot x_{F} - \dot x_{L}$)\\
		 &                              & \small$TTC$ - ($x_{L}$ - $x_{F}$ - $l_V$) = 0\\
\hline
\end{tabular}
\caption{\label{ATTCform} \textbf{Definition of different TTC types} \cite{BAMY} \cite{ref2}}
\end{table}

ATTC is based on fundamental kinematic equations, and solving these equations provides computable expressions for the TTC value. ATTC is divided into three different TTC types. $TTC_1$ represents the normal TTC calculation, $TTC_2$ represents MTTC, and $TTC_3$ can be interpreted as an extension of MTTC. $TTC_3$ also accommodates linear changes in acceleration and deceleration over time. However, it should be noted that not all acceleration or deceleration phases of vehicles are linear, as movement can be uneven. The implementation of ATTC relies on determining which assumption is currently fulfilled. \cite{BAMY} \cite{ref2}
 
\subsection{Time Headway (THW)}

THW measures the time that elapses between two vehicles reaching the same position \cite{BAMY} \cite{ref4}:

\begin{equation}
THW = \frac{x_{L} - x_{F} - l_V}{v_{F}}
\end{equation}

THW is similar to TTC, but it is defined solely by the distance between the following and leading vehicles and the speed of the following vehicle. It does not take into account the dynamics of the leading vehicle. Consequently, THW does not consider the influence of the leading vehicle, which can be significant in certain driving scenarios. \cite{BAMY} \cite{ref5}

\newpage
\subsection{Time To Stop (TTS)}
TTS \cite{ref6} considers the time needed to bring a vehicle to a complete stop under different environmental conditions. It can be seen as a correction of TTC, incorporating factors such as road and weather conditions through the friction coefficient $\mu$. TTS is based on three different deceleration values $a_i \in \{D, A, G\}$, representing different levels of criticality (D: Dangerous, A: Attentive, G: Gentle). A Gaussian model is employed to determine threat scores $\varphi_i$ for each level, accounting for deviations from TTC due to $\mu$, different deceleration profiles, and the uncertainty $\sigma$ of the Gaussian model. However, TTS still relies on TTC as an underlying assumption.

The threat scores $\varphi_i$ for the different levels are defined as follows:

\begin{align}
\varphi_D(\tau = D\,|\,TTS_{a_D},TTC)=\begin{cases}
			1, & TTC \leq TTS_{a_D} \\
            e^{-\frac{\Delta t_{D}^{2}}{2 \sigma^2}}, & \text{Otherwise}
		 \end{cases}
\end{align}
\begin{align}
\varphi_A(\tau = A\,|\,TTS_{a_A},TTC) &= e^{-\frac{\Delta t_{A}^{2}}{2 \sigma^2}}
\end{align}
\begin{align}
\varphi_G(\tau = G\,|\,TTS_{a_G},TTC)=\begin{cases}
			e^{-\frac{\Delta t_{G}^{2}}{2 \sigma^2}}, & TTC \leq TTS_{a_G} \\
            1, & \text{Otherwise}
		 \end{cases}
\end{align}

where:
\begin{align}
TTS_{a_i}(\tau = i,v_F) &= \mu \frac{v_F}{a_i}, \; i \in \{D, A, G\} \\
\Delta t_i &= TTC-TTS_{a_i}, \; i \in \{D, A, G\}
\end{align}

The probability $P_i$ for the different criticality levels can be calculated based on the threat scores:

\begin{align}
 P(\tau = i\,|\,TTS_{a_i},TTC) = \frac{\varphi_i(\tau = i\,|\,TTS_{a_i},TTC)}{\sum_{i} \varphi_i(\tau = i\,|\,TTS_{a_i},TTC)}, \; i \in \{D, A, G\}
\end{align}

For binary classification of safety-critical versus non-safety-critical driving scenarios, the probability calculation can be simplified as follows:

\begin{align}
 P_D(\tau = i\,|\,TTS_{a_i},TTC) = \frac{\varphi_D(\tau = D\,|\,TTS_{a_D},TTC)}{\sum_{i} \varphi_i(\tau = i\,|\,TTS_{a_i},TTC)}, \; i \in \{D, A, G\}
\end{align}

Based on $P_D$, a distinct split between safety-critical and non-safety-critical driving scenarios can be defined:

\begin{align}
b_{Crit} = \left\{
\begin{array}{ll}
1,& P_D \geq P_{D,thres} \\
0, & \, \textrm{Otherwise}
\end{array}
\right.    
\end{align}

where $P_{D,thres}$ represents the probability threshold for interpreting driving scenarios as safety-critical ($b_{Crit} = 1$) or non-safety-critical ($b_{Crit} = 0$). For more information on TTS, refer to the publication \cite{ref6}.

\subsection{Adaptive Difference Space Stopping (ADSS)}

ADSS is a novel safety indicator considered for the underlying use case. It is based on the Difference Space Stopping (DSS) metric \cite{DSS}. In general, both metrics are defined by calculating the difference between the space distance and stopping distance of two vehicles following each other. The space distance is obtained by summing the braking distance $x_{B,L}$ of the leading vehicle and the effective distance $d_V$ between the leading and following vehicles. The stopping distance is calculated by summing the brake reaction distance $x_{R,F}$ and the braking distance $x_{B,F}$ of the following vehicle. Both metrics can be interpreted as the frozen position of both vehicles when the leading vehicle suddenly applies the brakes, causing the following vehicle to also brake.

The main difference between ADSS and DSS lies in the deceleration values applied to both vehicles. For DSS, the maximum deceleration value $a_{B,max} = \mu \, g$ is used for braking. As a result, DSS can be seen as a safety metric to determine whether a collision could be avoided in theory. In contrast, ADSS considers actual deceleration values ($a_{B,L}$, $a_{B,F}$) for each vehicle individually. Generally, these deceleration values are smaller or equal to the maximum achievable braking deceleration value ($a_{B,L} \leq a_{B,max}$, $a_{B,F} \leq a_{B,max}$), which depends on driver behavior, road conditions, and the coefficient of friction $\mu$.

Deceleration values and reaction times are influenced by psychological and physiological characteristics of the driver, as well as the strength and timing of the braking system activation. In practice, drivers do not react perfectly in terms of timing and adherence to theoretical expectations. Additionally, vehicles themselves have response delays and tolerances. Although autonomous cars aim to minimize these gaps, there will always be a minimal gap between driver reaction and the activation of the vehicle's braking system. By incorporating actual deceleration values and realistic reaction times, ADSS becomes a more generalized safety indicator, providing more reliable datasets for testing purposes in autonomous cars. Therefore, the use of actual deceleration values and realistic reaction times increases ADSS's potential to identify safety-critical driving scenarios.

ADSS is defined by the equation:
\begin{align}
ADSS = \left(d_V + x_{B,L}\right) - \left(x_{R,F} + x_{B,F}\right) = \left( \left( x_L-x_F-l_V \right) + \frac{v_{L}^2}{2 \, a_{B,L}} \right) - \left( v_F \, t_R + \frac{v_{F}^2}{2 \, a_{B,F}} \right)
\end{align}
where $a_{B,L} = \max(| a_L |,a_{B,max})$ and $a_{B,F} = \max(| a_F |,a_{B,max})$. The reaction time $t_R$ can be modeled as a time-shifted Gamma distribution as defined in \cite{pubS1}.

To ensure that safety indicator metrics like ADSS cover a wide range of driving scenarios, it is crucial to assess their coverage by considering various possible driving scenarios. One effective approach is to use matrices that focus on speed and acceleration or deceleration combinations of both vehicles (in the context of two vehicles following each other).

The following matrices (Table \ref{vMatrix} and Table \ref{aMatrix}) provide an example of such an assessment:

			\begin{table}[H]
			\centering
			\begin{minipage}{0.45\textwidth}		
	\begin{tabular}{|l|l|l|l|}
		\hline
		& \textbf{$v_L < 0$} & \textbf{$v_L = 0$} & \textbf{$v_L > 0$} \\ \hline
		\textbf{$v_F < 0$} & 0 & 0 & 0 \\ \hline
		\textbf{$v_F = 0$} & 0 & 0 & 0 \\ \hline
		\textbf{$v_F > 0$} & 0 & 1 & 1 \\ \hline
	\end{tabular}
			\caption{\label{vMatrix} $v$ Matrix (Safety-relevant)}
			\end{minipage}\hfill
			\begin{minipage}{0.45\textwidth}
	\begin{tabular}{|l|l|l|l|}
		\hline
		& \textbf{$a_L < 0$} & \textbf{$a_L = 0$} & \textbf{$a_L > 0$} \\ \hline
		\textbf{$a_F < 0$} & 1 & 0 & 0 \\ \hline
		\textbf{$a_F = 0$} & 1 & 0 & 0 \\ \hline
		\textbf{$a_F > 0$} & 0 & 0 & 0 \\ \hline
	\end{tabular}
			\caption{\label{aMatrix} $a$ Matrix (Safety-relevant)}
			\end{minipage}
			\end{table}

In these matrices, the rows and columns define the value ranges of speeds and accelerations of both vehicles. A cell value of 0 represents a non-safety-critical driving scenario, while a cell value of 1 represents a safety-critical driving scenario. It is worth noting that the matrices alongside cell values correspond to both potential rear-end collisions and are associated with DSS and ADSS. In summary, there are four combinations of speeds and accelerations with a cell value of 1. A good coverage for a safety indicator metric, such as ADSS, should include exactly those four combinations.

For ADSS, it covers the following driving scenarios (potential rear-end collisions):
\begin{itemize}
\item Vehicle in front moves forward or comes to a standstill while the vehicle behind moves forward.
\item Vehicle in front brakes while the vehicle behind either also brakes or continues to drive at constant speed.
\end{itemize}

Theoretical combinations in which the following vehicle accelerates are not considered. This assumption holds when the driver of the following vehicle accelerates only to increase speed (not to create a potential safety-critical driving scenario with a vehicle in front).

These considerations ensure that ADSS focuses on safety-critical driving scenarios.

\section{Comparison of safety indicator metrics}
When selecting a suitable safety indicator metric for a specific use case, it is important to compare different metrics against each other. To facilitate this comparison, corresponding evaluation criteria are defined and categorized in Table \ref{classEval}. \cite{BAMY} \cite{pubS2} \cite{pubS3}  

\begin{itemize}
	\item \textbf{Complexity (C)}: The complexity of a safety indicator metric refers to its implementation and time complexity. It depends on factors such as the number of data points and influences the runtime and efficiency of the algorithm used.
	\item \textbf{Applicability (A)}: Applicability assesses how easily a safety indicator metric can be applied. It considers the level of effort required for implementation.
	\item \textbf{Transparency (T)}: Transparency measures the clarity and availability of information about the safety indicator metric in literature. It indicates whether sufficient well-founded and detailed information is available.
	\item \textbf{Robustness (R)}: Robustness indicates how well a safety indicator metric can handle different driving scenarios and variations in driving data, such as changes in kinematic values over time.
	\item \textbf{Parametrizability (P)}: Parametrizability evaluates whether the safety indicator metric includes parameters and how easily they can be set.
	\item \textbf{Interpretability (I)}: Interpretability assesses how easily the results of the safety indicator metric can be interpreted. Metrics with standardized or limited ranges of values have higher interpretability, while others may require expert knowledge or empirical values for interpretation.
	\item \textbf{Effectiveness (E)}: Effectiveness represents the overall validation and quality representation of the safety indicator metric, taking into account all the other evaluation criteria.
\end{itemize}

\begin{table}[H]
	\centering
	\begin{tabular}{l|c|p{0.69\linewidth}}
		\hline
		Complexity & + & Safety indicator metric has low time complexity and run time.\\

		& o & Safety indicator metric has moderate time complexity and run time.\\

		& - & Safety indicator metric has high time complexity and run time.\\
		\hline
		Applicability & + & Safety indicator metric is relatively easy to implement, or a reference implementation is available.\\

		& o & Safety indicator metric can only be implemented to a limited extent and no reference implementation is available.\\

		& - & Safety indicator metric is fundamentally difficult to implement and no reference implementation is available.\\
		\hline
		Transparency & + & Safety indicator metric is fully transparently described and understandable.\\

		& o & Safety indicator metric is moderately transparently described.\\

		& - & Safety indicator metric requires in-depth mathematical knowledge, or its description is incomplete.\\
		\hline
		Robustness & + & Safety indicator metric is robust against different driving scenarios and data. Assumptions of the safety indicator metric can be easily maintained. \\
		
		& o & Safety indicator metric is moderately robust against different driving scenarios and data. Assumptions of the safety indicator metric can be moderately maintained.  \\

		& - & Safety indicator metric is not robust against different driving scenarios and data. Assumptions of the safety indicator metric can be difficult to maintain. \\
		\hline
		Parametrizability & + & Safety indicator metric doesn't include parameters or includes only a few easily adjustable parameters.\\

		& o & Safety indicator metric includes several parameters, not all of them are easily adjustable.\\

		& - & Safety indicator metric includes several or many parameters, and all or most of them are not easily adjustable.\\
		\hline
		Interpretability & + & The result of the safety indicator metric is easy to interpret.\\

		& o & The result of the safety indicator metric is moderately interpretable.\\

		& - & The result of the safety indicator metric is difficult to interpret.\\
		\hline
	\end{tabular}
	\caption{Classification of individual Evaluation Criteria}
	\label{classEval}
\end{table}

\begin{itemize}
\item The sign "+" represents a positive characteristic, indicating that the safety indicator metric performs well in that particular evaluation criterion.
\item The sign "o" represents a neutral characteristic, indicating that the safety indicator metric has a moderate performance in that criterion.
\item The sign "-" represents a negative characteristic, indicating that the safety indicator metric does not perform well in that specific criterion.
\end{itemize}

\begin{figure}[H]
\centering
    \includegraphics[scale=0.56]{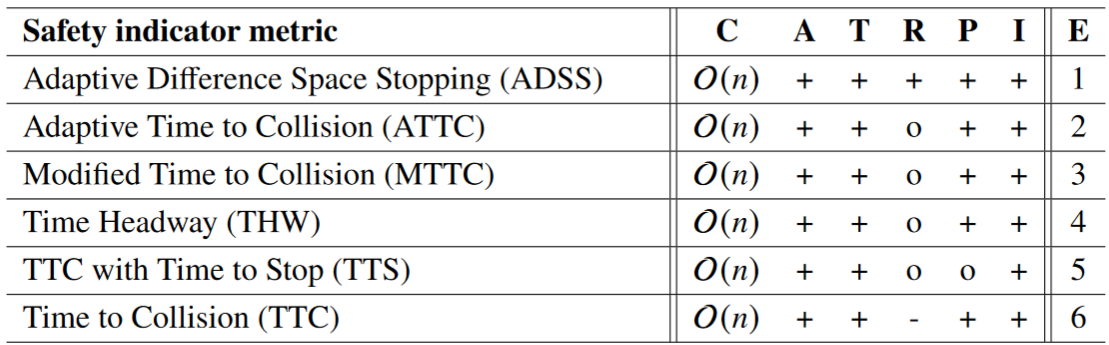}
    \caption{Comparison of Safety Indicators according to Evaluation Criterion}
    \label{tab:haupt:eval}
\end{figure}

Using this interpretation, the safety indicator metrics can be compared based on the evaluation criteria as shown in Figure \ref{tab:haupt:eval}.

Based on the additional information, the comparison in Table \ref{tab:haupt:eval} can be described further:
\begin{itemize}
\item ADSS stands out as the most suitable safety indicator metric, as it fulfills all the expected requirements and has positive characteristics across all evaluation criteria.
\item ATTC, MTTC, TTS, and THW exhibit a moderate level of robustness against different driving scenarios and data. However, these metrics are constrained by assumptions that may not be maintained in real-world applications. \cite{BAMY}
\item ATTC is ranked higher than MTTC because it also considers linear changes in accelerations for both vehicles, providing a more comprehensive analysis. \cite{BAMY}
\item MTTC is ranked higher than THW because it takes into account the dynamics of the leading vehicle and is dependent on the relative acceleration between the two vehicles. \cite{BAMY}
\item TTS utilizes parameters that are difficult to specify. Additionally, TTS is based on TTC and is therefore subject to the limitations of TTC itself. \cite{BAMY}
\end{itemize}

\newpage 

\section{Validation}
Validating the effectiveness of the ADSS is essential to assess its feasibility. Therefore, a pipeline has been developed, comprising an input unit, a processing unit, and an output unit. The following sections provide a detailed description of each unit.

\subsection{Input Unit}
\begin{figure}[H]
\centering
\includegraphics[scale=0.7]{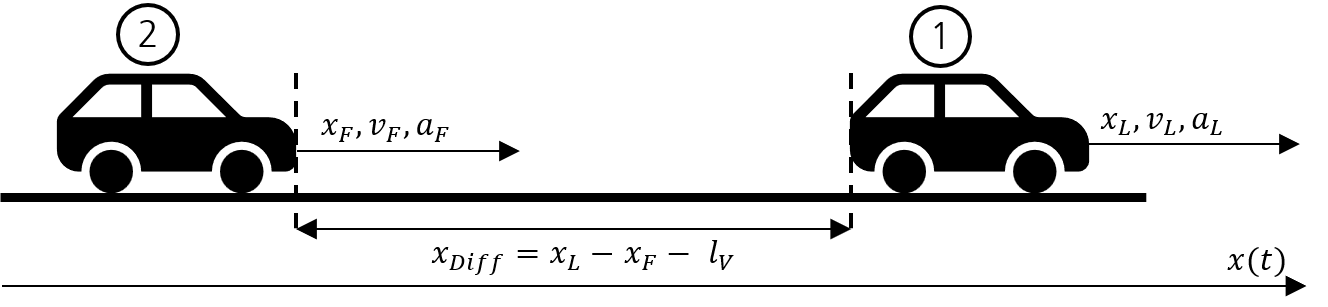}
\caption{Validation: Follow-Up Drive}
\label{ValidScene}
\end{figure}

To effectively demonstrate the capabilities of ADSS, it was crucial to establish a decision criterion for selecting an appropriate driving scenario as a baseline. The chosen driving scenario should possess a relatively high frequency in real-world practice and be easily modeled. Based on these criteria, the decision was made to simulate a follow-up drive, where two vehicles drive in a straight line with one vehicle following the other. The specifics of this driving scenario are described in more detail below. Figure \ref{ValidScene} illustrates the arrangement of the follow-up drive.

The driving scenario involves two vehicles moving in the longitudinal direction, where Vehicle 2 follows Vehicle 1 (follow-up drive). Both vehicles execute braking maneuvers at specific times. In this scenario, the primary kinematic variables under consideration are the position ($x$), speed ($v$), acceleration ($a$), and time ($t$). While the jerk ($j = \dot{a}$) theoretically holds significance, the focus of this analysis lies in the overall macroscopic motion of the vehicles rather than microscopic details. Hence, measures such as jerk damping are considered as implemented, enabling the neglect of jerk-related considerations.

A mathematical model, built upon the insights are presented in publication \cite{pubS1}, was utilized to generate synthetic data in the form of time series. This data was employed to simulate the movement of the vehicles within the driving scenario. The modeling process entailed configuring several parameters and initial conditions.

The initial distance between the vehicles was defined as $d_0 = x_{0,L} - x_{0,F} - l_V = 15.4$ meters, where $l_V = 4.6$ meters \cite{VehLen} corresponds to the length of the vehicle. This initial distance accounts for the spatial separation between the leading vehicle (Vehicle 1) and the following vehicle (Vehicle 2) at the beginning of the driving scenario.

To incorporate variations in the vehicles' movements, publication \cite{pubS1} introduces different initial speeds ($v_0$), initial accelerations ($a_0$), and reaction times ($t_R$). Probability distributions are utilized to model these variations, with uniform distributions specifically employed for both initial speeds and initial accelerations, whereas the reaction times ($t_R$) are modeled using a gamma distribution.

\newpage
The following initial speeds are assigned to the vehicles:

\begin{align}
v_{0,L} &= v_{0,m,L} + v_{0,var,L} = 22.\overline{22} \, \frac{m}{s} + v_{0,var, L} \in \{-1, -0.95, \ldots, 0.95, 1\} \, \frac{m}{s} \\
v_{0,F} &= v_{0,m,F} + v_{0,var,F} = 25 \, \frac{m}{s} + v_{0,var,F} \in \{-1, -0.95, \ldots, 0.95, 1\} \, \frac{m}{s}
\end{align}

Vehicle 2 maintains an average speed of 90 km/h, which is 10 km/h faster than the average speed of Vehicle 1, set at 80 km/h. This selection of speeds aims to create a higher occurrence of safety-critical driving scenarios. The initial accelerations for both vehicles are defined as follows:

\begin{align}
a_0 = a_{0,m} + a_{0,var} = 7 \frac{m}{s^2} + a_{0,var} \in \{-1, -0.95, \ldots, 0.95, 1\} \frac{m}{s^2}
\end{align}

It is stated that a reaction time of $t_R = 0.7$ seconds corresponds to the expected value of the distribution function \cite{pubS1}. Hence, this value is utilized for the validation process, ensuring that the simulated reaction times align with the expected average reaction time \cite{pubS1}.

In the validation process, a total of 1000 time series were considered, with each time series comprising 10 data points. The time vector, denoted as $t=\left[0, 0.25, \ldots, 2.0, 2.25\right]$ seconds, was defined with a step size of $\Delta t = 0.25$ seconds. Each individual time series represents a follow-up drive of both vehicles, encompassing all the relevant kinematic variables such as position ($x$), speed ($v$), acceleration ($a$), and time ($t$). These time series encompass the initial values as well as the varying values previously mentioned, allowing for a comprehensive analysis of the driving scenario.

\subsection{Processing Unit}
The processing unit consists of two main components: the calculation of the ADSS  value and the classification of the safety relevance of the driving scenario based on the ADSS value.

The ADSS value is calculated for each data point in each time series. Each time series represents either a safety-critical driving scenario or a non-safety-critical driving scenario. A driving scenario is classified as safety-critical if and only if at least one data point in the corresponding time series yields a safety-critical ADSS value. The classification of safety relevance based on the ADSS can be defined as follows:

\begin{align}
	b_{Crit} = \left\{
	\begin{array}{ll}
		1 & \forall \, ADSS \leq 0 \land (a_L < 0 \land a_F < 0) \\
		0 & \, \textrm{Otherwise} \\
	\end{array}
	\right.
\end{align}

Hence, a driving scenario can be classified as safety-critical ($b_{Crit} = 1$) if both vehicles are braking at current time ($a_L < 0 \land a_F < 0$) and the calculated ADSS value is less than or equal to 0 meters. Otherwise, the driving scenario is considered as non-safety-critical ($b_{Crit} = 0$) at current time. This ADSS-based evaluation is performed for each data point in the time series where both vehicles are braking.

\newpage

\subsection{Output Unit}
\begin{figure}[H]
\centering
\includegraphics[scale=0.70]{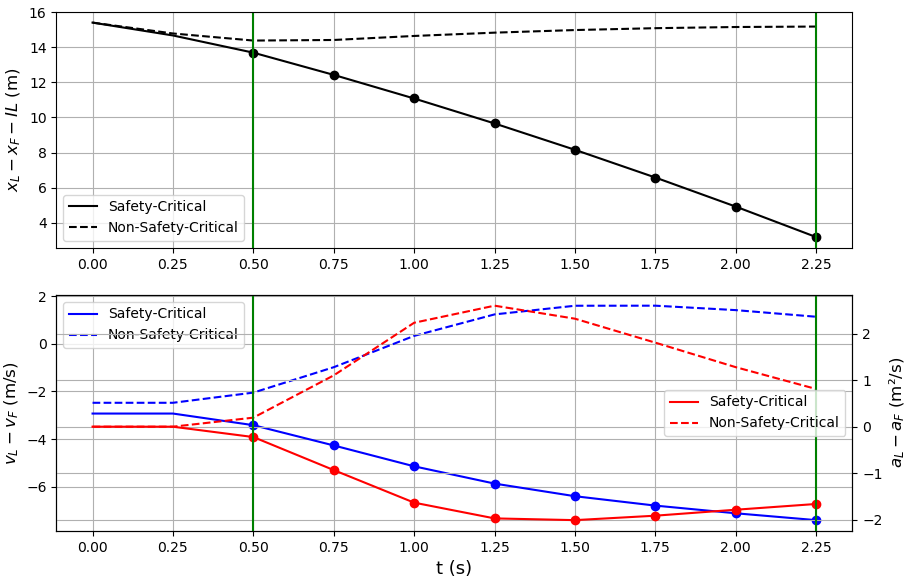}
\caption{Validation: Criticality of Follow-Up Drive (Time Series Data)}
\label{Valid1}
\end{figure}

The output unit is responsible for visualizing the time series to showcase the driving scenarios considered during the validation process. The time series are presented as a 2D representation over time, highlighting the differences in position, speed, and acceleration between the two vehicles. Figure \ref{Valid1} illustrates this visualization approach.

Figure \ref{Valid1} consists of two subplots stacked on top of each other. The upper subplot displays the distance between the two vehicles over time, while the lower subplot depicts the differences in speed and acceleration between the vehicles over time.

In general, two time series are considered, and they have been classified in terms of their safety relevance based on the ADSS. Solid lines represent a safety-critical driving scenario, while dashed lines represent a non-safety-critical driving scenario. The green area between 0.5 and 2.25 seconds visualizes the time range in which both vehicles are braking and classified as safety-critical according to ADSS. For the time series classified as non-safety-critical, the difference in acceleration is always positive, indicating that both vehicles tend to move away from each other during this highlighted time period in green. This is further emphasized by the increased difference in speed between the vehicles during the same time period.

For the time series classified as safety-critical, the differences in acceleration and speed become smaller (greater negative values), resulting in a constant decrease in the distance between the vehicles (indicating an increasing potential for a rear-end collision). It is important to note that the reaction time is modeled using the expected value (0.7 seconds) of the underlying distribution.

\section{Conclusion}
Validating and testing autonomous cars across various driving scenarios is crucial to establish release criteria and ensure human acceptance. Different methods exist for categorizing driving scenarios as either safety-critical or non-safety-critical. In this context, the Adaptive Difference Space Stopping (ADSS) metric proves to be effective in providing comprehensive and reliable coverage, particularly for longitudinal driving scenarios involving two vehicles following each other. The movement of both vehicles is defined using time series data that encapsulate the fundamental kinematic equations of motion.

\end{document}